\documentclass[12pt]{article}
 \usepackage[utf8]{inputenc}

 \usepackage{amsmath}
 \usepackage{amsfonts}
\usepackage{url}
\usepackage{xcolor}
\usepackage[english]{babel}

\begin{document}
%--------------------------------------------------------------------------

\begin{center}
{\Large
	{\sc Consistance de la minimisation du risque empirique pour l'optimisation de l'erreur relative moyenne}
}
\bigskip

 Arnaud de Myttenaere$^{1}$ \& B\'en\'edicte Le Grand$^{2}$ \& Fabrice Rossi$^{3}$
\bigskip

{\it
$^{1}$ Viadeo, ademyttenaere@viadeoteam.com 

$^{2}$ Université Paris 1 Panthéon Sorbonne CRI, benedicte.le-grand@univ-paris1.fr

$^{3}$ Université Paris 1 Panthéon Sorbonne SAMM, fabrice.rossi@univ-paris1.fr
}
\end{center}
\bigskip
%--------------------------------------------------------------------------

{\bf R\'esum\'e.} Nous nous intéressons au problème de la minimisation de l'erreur relative moyenne dans le cadre des modèles de régression. Nous montrons que l'optimisation de ce critère est équivalente à la minimisation de l'erreur absolue par régressions pondérées et que l'approche par minimisation du risque empirique est, sous certaines hypothèses, consistante pour la minimisation de ce critère.
\smallskip

{\bf Mots-cl\'es.} Optimisation, MAPE, consistance
\bigskip\bigskip

{\bf Abstract.} We study in this paper the consequences of using the Mean
Absolute Percentage Error (MAPE) as a measure of quality for regression
models. We show that finding the best model under the MAPE is equivalent
to doing weighted Mean Absolute Error (MAE) regression. We also show that, under some asumptions,
universal consistency of Empirical Risk Minimization remains possible
using the MAPE.
\smallskip


%--------------------------------------------------------------------------

%%%
%%% Introduction
%%%
\section{Introduction}\label{sec:introduction}
Nous étudions le cadre classique des régressions, en supposant des couples d'observations $Z = (X,Y)$, à valeurs dans $\mathcal{X}\times\mathbb{R}$ où $\mathcal{X}$ est un espace muni d'une métrique. 
La qualité d'un modèle $g$ (fonction définie sur $\mathcal{X}$ et à valeurs dans $\mathbb{R}$) est mesurée à partir d'une fonction de perte $l$, qui est classiquement les moindres carrés (MSE: \emph{Mean Square Error}), l'erreur absolue (MAE: \emph{Mean Absolute Error}), ou l'erreur relative moyenne (MAPE: \emph{Mean Absolute Percentage Error}):
\[ l_{MAPE}(p, y) = \bigg| \frac{p-y}{y} \bigg| \]

avec les conventions que pour tout $a\neq 0$, $\frac{a}{0}=\infty$ et $\frac{0}{0}=1$.
Le risque d'un prédicteur $g$ est défini comme l'espérance de la perte: $L_l(g)=\mathbb{E}(l(g(X),Y))$. Le risque empirique est alors la moyenne empirique de la fonction de perte calculée sur l'ensemble d'apprentissage:
\begin{equation}\label{eq:empirical:risk}
\widehat{L}_l(g)_N=\frac{1}{N}\sum_{i=1}^Nl(g(X_i),Y_i).
\end{equation}

L'enjeu pratique consiste à déterminer comment minimiser $L_{MAPE}(g)$. D'un point de vue théorique, nous nous sommes intéressés à la consistance de la méthode de minimisation du risque empirique (ERM) dans le cas de la MAPE.

%%%
%%% Résolution pratique
%%%
\section{Résolution pratique}\label{sec:practical}
%\subsection{Optimisation}
D'un point de vue pratique, le problème consiste à minimiser $\widehat{L}_{MAPE}(g)_N$ sur une classe de modèles $G_N$, ce qui revient à résoudre:
\[
\widehat{g}_{MAPE,N}=\arg\min_{g\in G_N}\frac{1}{N}\sum_{i=1}^N\frac{|g(X_i)-Y_i|}{|Y_i|}.
\]

En considérant les quotients $\frac{1}{|Y_i|}$ comme des poids, ce problème peut être vu comme un cas particulier de la régression médiane (elle-même étant un cas particulier des régressions quantiles) qui minimise l'erreur absolue. Par conséquent, toute implémentation des régressions quantiles permettant l'utilisation de pondérations peut être utilisée pour trouver le modèle linéaire minimisant la MAPE. C'est par exemple le cas de la librairie R \texttt{quantreg} \cite{Koenker13}.

%%%
%%% Theoretical issues
%%%
\section{Considérations théoriques}\label{sec:theory}
D'un point de  vue théorique, nous nous sommes intéressés à la consistance des stratégies d'apprentissage lorsque la fonction de perte est la MAPE. Plus précisément, pour une fonction de perte $l$, nous définissons $L^*_l=\inf_{g}L_l(g)$, où le minimum est calculé sur l'ensemble des fonctions mesurables de $\mathcal{X}$ dans $\mathbb{R}$ et notons $L^*_{l,G}=\inf_{g\in G}L_l(g)$, où $G$ est une classe de modèles. 

Dans ce travail, nous nous intéressons à la méthode de minimisation du risque empirique (ERM), pour lequel $\widehat{g}_{l,N}=\arg\min_{g\in G_N}\widehat{L}_l(g)_N$, et nous montrons la consistance de l'ERM dans le cas de la MAPE:

\[ \forall \epsilon > 0, \mathbb{P}\left\{ lim_{n \to +\infty}\left(\sup_{g\in G_N}\left|\widehat{L}_{mape}(\widehat{g}_{l,N})_N-L_{mape}(g)\right| \right) > \epsilon \right\} = 0 \]

Ce résultat a déjà été établi pour certaines fonctions de pertes (MSE et MAE par exemple), mais ne peut être généralisé au cas de la MAPE car les propriétés nécessitent l'hypothèse de continuité uniforme au sens de Lipschitz (voir par exemple le lemme 17.6 dans \cite{ab-nnltf-99}), qui n'est pas vérifiée dans le cas de la MAPE.

La preuve proposée s'effectue en deux étapes. D'abord nous montrerons qu'il est possible de borner la probabilité à contrôler par une quantité dépendant du $L_p$ convering number de la classe de modèles considérée. Puis, en contrôlant ce dernier par la VC-dimension de la classe de fonction considérée, nous verrons qu'il est possible, sous certaines hypothèses, d'assurer la consistance de la méthode de minimisation du risque empirique.

%%% $L_p$ convering numbers
\subsection{$L_p$ convering numbers}

Etant donnée une classe de modèles $G_N$ et une fonction de perte $l$, nous noterons
\[ H(G_N,l)=\{h: \mathcal{X}\times \mathbb{R}\rightarrow \mathbb{R}^+,h(x,y)=l(g(x),y)\ |\ g\in G_N\},\]
et 
\[ H^+(G_N,l)=\{h: \mathcal{X}\times \mathbb{R}\times \mathbb{R}\rightarrow \mathbb{R}^+,\
h(x,y,t)=\mathbb{I}_{t\leq l(g(x),y)}\ |\ g\in G_N\}.
\]
Lorsqu'il n'y a pas d'ambiguité compte tenu du contexte, nous simplifierons ces notations par $H_{N,MAPE}$ pour $l=l_{MAPE}$ et la classe $G_N$ considérée.

Pour tout $\epsilon>0$, une $\epsilon_p$-couverture de taille $p$ d'une classe de fonctions $\mathcal{F}$ définies sur $\mathcal{Z}$ et à valeurs dans $\mathbb{R}^+$ est une collection finie $f_1, \dots, f_p$ de $\mathcal{F}$ telle que $\min_{1\leq i\leq p}\|f-f_i\|_{p,D}<\epsilon$ pour tout $f \in \mathcal{F}$ et un jeu de donné D observé, où $\|.\|_p$ désigne la norme $L_p$.

Le $L_p$ covering number correspond au nombre de $\epsilon_p$-couverture de $\mathcal{F}$ et est noté $\mathcal{N}_p(\epsilon, \mathcal{F}, D)$. Comme pour tout $h_1$, $h_2$ dans $H_{n,MAPE}$ la quantité définie par 
\[ \| h_1 - h_2\|_\infty = \sup_{(x,y) \in \mathcal{X}\times \mathbb{R}} \frac{| |g_1(x) - y| - |g_2(x) - y| |}{|y|}\]

n'est en général pas bornée lorsque $y$ tend vers 0, les résultats liés à la MSE et MAE ne sont pas applicables directement. Dans la suite nous supposerons donc qu'il existe $\lambda > 0$ tel que $|Y| \geq \lambda$

En supposant que pour tout $g \in G_N$, $\| g \|_{\infty} \leq B_{G_N}$, et $|Y|\geq \lambda$, on peut montrer que $B_{H(G_N, l_{MAPE})} = 1 + \frac{B_{G_N}}{\lambda}$ et le théorème 9.1 de \cite{gyorfi_etal_DFTNR2002} donne (avec $B_{N,l}=B_{H(G_N,l)}$):

\begin{equation}\label{eq:ULLN}
P\left\{\sup_{g\in G_N}\left|\widehat{L}_{l}(g)_N-L_{l}(g)\right|>\epsilon\right\}
\leq 8\mathbb{E}\left(\mathcal{N}_{p}\left(\frac{\epsilon}{8},H(G_N,l),D\right)\right)e^{-\frac{N\epsilon^2}{128B_{N,l}^2}}.
\end{equation}

Ce qui permet d'avoir une borne de la quantité à contrôler.

%%% VC-dimension
\subsection{VC-dimension}
Une façon de borner les covering numbers consiste à utiliser les VC-dimensions. On peut montrer $k$ points sont séparés par $H^+(G_N, l_{MAE})$ si, et seulement si, ils sont également séparés par $H^+(G_N, l_{MAPE})$. En d'autres termes, la VC-dim de la classe de fonction considérée est inchangée selon  que la fonction de perte est la MAE ou la MAPE. D'après le théorème 9.4 de \cite{gyorfi_etal_DFTNR2002}, si $V_{l}=VC_{dim}(H^+(G_N,l))\geq 2$, $p\geq 1$, et $0<\epsilon<\frac{B_{N,l}}{4}$, on a alors
\begin{equation}\label{eq:covering:VC}
\mathcal{N}_{p}(\epsilon,H(G_N,l),D)\leq 3\left(\frac{2eB_{N,l}^p}{\epsilon^p}\log \frac{3eB_{N,l}^p}{\epsilon^p}\right)^{V_{l}}.
\end{equation}

\subsection{Consistance}
La consistance de l'algorithme ERM dans le cas de la MAPE peut être montrée de façon similaire au théorème 10.1 de \cite{gyorfi_etal_DFTNR2002}. Supposons donnée une série de classe de modèles, $(G_n)_{n \geq 1}$ telle que $\cup_{n \geq 1} G_n$ est dense dans l'ensemble des fonctions mesurables de $\mathbb{R}^p$ dans $\mathbb{R}$ selon la norme $L^1(\mu)$ pour toute mesure $\mu$. Supposons en outre les $G_n$ uniformément bornés par $B_{G_n}$ et de VC-dim finie $V_n = VCdim(H^+(G_n, l_{MAPE}))$. Remarquons que, pour que ces conditions restent compatibles avec l'hypothèse de densité il est nécessaire que $\lim_{n \to \infty} v_n = \infty$ et $\lim_{n \to \infty} B_{G_n} = \infty$.

Alors, en supposant que $|Y| \geq \lambda$ (presque surement) et que $\lim_{n\rightarrow \infty}\frac{v_nB_{G_n}^2\log B_{G_n}}{n}=0$, on obtient à partir des équations \ref{eq:ULLN} et \ref{eq:covering:VC}:

\[
P\left\{\sup_{g\in G_N}\left|\widehat{L}_{l}(g)_N-L_{l}(g)\right|>\epsilon\right\} \leq K(n,\epsilon),
\]

avec
\[
K(n,\epsilon) = 24\left( \frac{16eB_n}{\epsilon}\log \frac{24eB_n}{\epsilon}\right)^{v_n}e^{-\frac{n\epsilon^2}{128 B_n^2}} \quad \mbox{ et } \quad B_n = 1 + \frac{B_{G_n}}{\lambda}.
\]

Les hypothèses garantissent $\sum_{n\geq 1} K(n, \epsilon) < \infty$ pour tout $\epsilon > 0$, ce qui assure la convergence presque sure de $L_{MAPE}(\widehat{g}_{l_{MAPE},n})$ vers $L^*_{MAPE}$

%--------------------------------------------------------------------------

%\section*{Bibliographie}
\bibliographystyle{abbrv}
\bibliography{biblio}

\begin{thebibliography}{1}

\bibitem{ab-nnltf-99}
M.~Anthony and P.~L. Bartlett.
\newblock {\em Neural Network Learning: Theoretical Foundations}.
\newblock Cambridge University Press, 1999.

\bibitem{gyorfi_etal_DFTNR2002}
L.~Gy\"orfi, M.~Kohler, A.~Krzy\.zak, and H.~Walk.
\newblock {\em A Distribution-Free Theory of Nonparametric Regression}.
\newblock Springer, New York, 2002.

\bibitem{Koenker13}
R.~Koenker.
\newblock {\em quantreg: Quantile Regression}.
\newblock R package version 5.05., 2013.

\end{thebibliography}

\end{document}